\documentclass{article}

 \usepackage[preprint,nonatbib]{neurips_2019}

\usepackage[utf8]{inputenc} 
\usepackage[T1]{fontenc}    
\usepackage{hyperref}       
\usepackage{url}            
\usepackage{booktabs}       
\usepackage{amsfonts}       
\usepackage{nicefrac}       
\usepackage{microtype}      
\usepackage{times}
\usepackage{soul}
\usepackage{url}
\usepackage[utf8]{inputenc}
\usepackage[small]{caption}
\usepackage{graphicx}
\usepackage{amsmath}
\usepackage{booktabs}
\usepackage{enumitem}
\usepackage{wrapfig}

\renewcommand{\vec}[1]{\mathbf{\MakeLowercase{#1}}}

\newcommand{\set}[1]{\boldsymbol{#1}}

\newcommand{\policy}{\pi}
\newcommand{\detpol}{\mu}
\newcommand{\expectation}{\mathbb{E}}
\newcommand{\param}{\theta}
\newcommand{\objective}{J}
\newcommand{\reward}{r}

\newcommand{\action}{a}

\newcommand{\state}{s}
\newcommand{\replayBuffer}{\set{D}}
\newcommand{\loss}{\mathcal{L}}
\newcommand{\decay}{\gamma}
\newcommand{\stateSet}{\set{S}}
\newcommand{\actionSet}{\set{A}}
\newcommand{\observationSet}{\set{O}}

\newcommand{\policySet}{\mathbf{\pi}}

\newcommand{\realnum}{\mathbb{R}}
\newcommand{\transfun}{T}
\newcommand{\transProb}{P}

\newcommand{\fullobservation}{\vec{x}}
\newcommand{\observation}{o}

\usepackage[acronym,style=alttree, toc=true, shortcuts, xindy, nomain, nonumberlist]{glossaries}
\usepackage{setspace}
\usepackage{filecontents}
\usepackage{silence}
\usepackage{dsfont}
\usepackage{pgfplots}
\usepackage{caption}
\usepackage{tikz}

\usepackage{pdfpages}
\usepackage{tabulary}
\usepackage{bm}
\usepackage[bottom]{footmisc}
\usepackage{amsfonts}

\graphicspath{{pics/}}

\usepgfplotslibrary{groupplots}

\pgfplotsset{every tick label/.append style={font=\small}}

\usepackage{algpseudocode}
\usepackage[ruled,vlined]{algorithm2e}

\SetEndCharOfAlgoLine{}
\newlength{\commentWidth}
\SetKwComment{Comment}{$\triangleright$\ }{}

\SetCommentSty{mycommfont}

\DontPrintSemicolon
\SetArgSty{textrm}
\SetKwProg{Fn}{Function}{:}{} 

\newcommand*\Let[2]{#1 $\gets$ #2}  

\makeatletter
\def\BState{\State\hskip-\ALG@thistlm}
\makeatother

\newglossary{symbols}{sym}{sbl}{List of Symbols}
\newglossary{notation}{not}{nt}{Notation}
\glssetwidest{THISWIDE} 
\makeglossaries
\loadglsentries{gloss.aux}
\makeindex

\title{Reducing Overestimation Bias in Multi-Agent Domains Using Double Centralized Critics}

\author{%
  Johannes Ackermann \thanks{Work performed while at the University of Tokyo} \\
  Technical University of Munich\\
  \texttt{johannes.ackermann@tum.de} \\
   \And
   Volker Gabler \\
  Technical University of Munich\\
  \texttt{v.gabler@tum.de} \\
   \AND
   Takayuki Osa \\
   Kyushu Institute of Technology, RIKEN\\
  \texttt{osa@brain.kyutech.ac.jp} \\
   \And
   Masashi Sugiyama \\
   RIKEN, The University of Tokyo \\
  \texttt{sugi@k.u-toyo.ac.jp}
}

\begin{document}

\maketitle

\begin{abstract}
	Many real world tasks require multiple agents to work together.
	Multi-agent reinforcement learning (RL) methods have been proposed in recent years to solve these tasks, but current methods often fail to efficiently learn policies.
	We thus investigate the presence of a common weakness in single-agent RL, namely value function overestimation bias, in the multi-agent setting.
	Based on our findings, we propose an approach that reduces this bias by using double centralized critics.
	We evaluate it on six mixed cooperative-competitive tasks, showing a significant advantage over current methods.
	Finally, we investigate the application of multi-agent methods to high-dimensional robotic tasks and show that our approach can be used to learn decentralized policies in this domain.
\end{abstract}

\section{Introduction}

In recent years, many real world problem settings have been modeled as multi-agent systems, be it smart-grid applications \cite{Li2012}, package routing \cite{Ye2015}, or road transportation \cite{Adler2002}.

While it is possible to regard these problems as a centralized single agent, with a large state and action space, and apply methods from single-agent \gls*{RL}, this leads to an action space that increases exponentially with the number of agents \cite{Mehta2005}.
Another approach is to assume independent learners \cite{Tan1993}, in which agents regard the influence of other agents as part of the environment.
However, due to the behavior of other agents changing over time, 
the transition probabilities change, 
leading to the Markov assumption being violated.
Therefore, recent research has focused on decomposing these systems into individual, decentralized agents during execution, while updating them in a centralized training phase, allowing to maintain the Markov property during training \cite{Oliehoek2008}.
Although recent research has introduced powerful \gls*{MARL} techniques based on this principle, such as counterfactual multi-agent (COMA) policy gradients \cite{Foerster2017}, or the \gls*{MADDPG} method \cite{Lowe2017}, their performance has not been studied as thoroughly as approaches in single-agent \gls*{RL}.
Motivated by findings in the single-agent case \cite{VanHasselt2015,Fujimoto2018}, which have shown it to generally suffer from an overestimation bias of the value function, we thus investigate this issue in \gls*{MARL} on the example of the popular \gls*{MADDPG} method.

Similarly to multi-agent tasks, complex robotic systems face the challenge of high-dimensional and continuous state-action spaces. 
A popular way to approach these tasks is decentralized control \cite{Ijspeert2008}, which requires a high amount of model knowledge.
Recently \cite{Sartoretti2018} presented a method to learn decentralized policies with \gls*{RL}, but it still relies on a shared or centralized meta-policy.
We propose a way to learn truly decentralized policies, by modeling robotic systems as multi-agent systems, eliminating the need for a centralized controller.

The main contribution of our work is a new method for \gls*{MARL}, that addresses overestimation bias and outperforms previous methods in most of the evaluated cooperative-competitive tasks.
Furthermore, we provide an approach to learn decentralized policies for high-dimensional robotic tasks, based on \gls*{MARL}.
We show that our method is able to learn decentralized policies on a simulated task and outperforms existing \gls*{MARL} methods.

\section{Background}
In this section, we explain the relevant background for our work, first explaining general methods in \gls*{RL}, then addressing more recent policy gradient algorithms.
\subsection{Markov Games}
In our work, we focus on Markov games \cite{Littman1994}, an extension of Markov decision processes to multi-agent domains.
A Markov game with $N$ agents consists of a state set $\stateSet$, a collection of action sets, $\actionSet_1,\ldots,\actionSet_N$, a transition function $\transfun : \stateSet \times \actionSet_1 \times \cdots \times \actionSet_N \to \mathrm{Dist}(\stateSet)$, with $\mathrm{Dist}(\stateSet)$ being a distribution over states.
Each agent has its own reward function $\reward_i : \stateSet \times \actionSet_1 \times \cdots \times \actionSet_N \to \realnum$, which depends on the actions of all agents.
Since we regard decentralized agents, they each posses a different observation set $\observationSet_i$, which is available to them during execution, and choose actions according to a policy $\policy_i : \observationSet_i \to \mathrm{Dist}(\actionSet_i)$.

The agents each aim to maximize their own total expected return $R_i = \sum_{t=0}^{t=T} \gamma^t r_i$, with a discount factor $0<\gamma\le1$ and time horizon $T$.
If $\reward_i =  k \reward_j$, $i\neq j$, the interaction is cooperative for $k>0$ and competitive for $k<0$	.

\subsection{Q-Learning}
Q-learning \cite{Sutton1998reinforcement} is an off-policy algorithm that learns the value of executing action $\action$ in state $\state$ in form of the expected return $Q^\policy(\state,\action) = \expectation[R|\state_t=\state,\action_t=\action]$, which can be recursively obtained as $Q^\policy(\state,\action) = \expectation_{\state'}[r(\state,\action) + \gamma \expectation_{\action'\sim\policy(\state')}[Q^\policy(\state',\action')]]$.
Assuming a greedy policy $\policy(\state) = \arg\max_\action(Q^\policy(\state,\action))$, Q-learning can be used to learn an optimal policy.

Mnih et al. \cite{Mnih2015} proposed an approach that approximates the Q-function with \glspl*{MLP}, called \gls*{DQN}. It uses a target network $Q^\policy_{\param'}$, whose parameters $\param'$ slowly follow the network parameters of $Q^\policy_{\param}$, to update the parameters of the $Q$-network.
Additionally, transitions $(\state,\action,\reward,\state')$ are stored in a replay buffer $\replayBuffer$.

Double Q-learning \cite{VanHasselt2010} found that Q-learning often overestimates the $Q$-value in stochastic environments, leading to a failure to learn an efficient policy.
To remove this positive bias, they proposed to learn two Q-functions $Q_1,Q_2$, which are updated using the value of the respective other function. For $Q_1$, this resolves to $Q_1(\state,\action) = Q_1(\state,\action) + \alpha(\reward + \gamma Q_2(\state',\action') - Q_1(\state,\action))$, with $\action' = \arg\max_\action Q_1(\state',\action)$. 

\subsection{Policy Gradient Methods}

Sutton et al. \cite{Sutton1999Gradient} took a different approach to optimizing the behavior of the agent, by directly using gradient descent on the parameters of the policy.
Their target $\objective(\param) = \expectation_{\state\sim p^\policy,\action\sim\policy_\param}[\sum_{t=0}^{\infty}\decay^t\reward_t]$ is defined as the expected total reward over a policy dependent state distribution $p^\policy$ and the gradient resolves to
\begin{equation}
\nabla_\param\objective(\param) = \expectation_{s\sim p^\policy,a\sim\policy_\param}\left[\nabla_\param\log\policy_\param(\action|\state)Q^\policy(\state,\action)\right]\,.
\label{eq:policyGradient}
\end{equation}

In 2014, Silver et al. \cite{Silver2014} derived a formulation of the policy gradient theorem for deterministic policies $\detpol : \stateSet \to \actionSet$ called \gls*{DPG}. They also showed that deterministic policies tend to learn significantly quicker than stochastic policies in some domains.

An algorithm for \gls*{RL} in continuous control problems, based on \gls*{DPG}, was presented in \cite{Lillicrap2016}, called the \gls*{DDPG}. It performs off-policy updates using transitions from a replay buffer $\replayBuffer$ and utilizes a target network, as in \gls*{DQN}.
Using this, the gradient in \eqref{eq:policyGradient} becomes
\begin{equation}
\nabla_\param\objective(\param) = \expectation_{s\sim D}
\left[\nabla_\param\detpol_\param(\state)\nabla_\action Q^\detpol(\state,\action)|_{\action=\detpol_\param(\state)}\right] \,.
\label{eq:DDPG}
\end{equation}
\subsection{TD3}
The \gls*{TD3} \cite{Fujimoto2018}  improves on \gls*{DDPG} by addressing the overestimation bias of the Q-function, similarly to double Q-learning.
They find that due to approximation errors of the \gls*{MLP}, combined with gradient descent, \gls*{DDPG} tends to overestimate the Q-value of state-action pairs, leading to a slower convergence.
\gls*{TD3} addresses this by using two Q-networks $Q_{\param_1},Q_{\param_2}$, along with two target networks. 
The Q-functions are updated with the target $y=r_t + \decay \min_{1,2} Q_{\param_i'}(\state',\action')$, while updating the policy with $Q_{\param_1}$.
Additionally, they introduce target policy smoothing by adding noise in the determination of the next action for the critic target $\action' = \detpol_{\param_\policy'}(\state') + \epsilon$, with $\epsilon$ being clipped Gaussian noise $\epsilon = \mathtt{clip} (\mathcal{N}(0,\sigma),-c,c)$, where $c$ is a tunable parameter.

Additionally they use delayed upolicy updates, and only update the policy $\pi$ and target network parameters $\theta_\policy', \theta_Q'$ once every $d$ critic updates.

\subsection{Multi-Agent Deep Deterministic Policy Gradient}
\label{bg:maddpg}
\gls*{MADDPG} \cite{Lowe2017} is an extension of \gls*{DDPG} to the multi-agent setting. 
It uses the decentralized execution with a centralized training setting, learning a centralized critic that has access to the policies of all agents.
This centralized Q-function, representing the expected future reward of agent $i$, is then learned with
\begin{equation}
Q_i^\policySet (\fullobservation, \action_1,...,\action_N) = 
\expectation_{\reward, \fullobservation'}
[\reward_i + \gamma Q'^\policySet_i(\fullobservation',\detpol_1(\observation_1'),...,\detpol_N(\observation_N') ]
\label{eq:maddpg:qupdate}
\end{equation}
Using this Q-function, the deterministic policy of agent $i$ can be optimized by gradient descent:
\begin{equation}
\begin{split}
\nabla_{\param_i}J(\detpol_i) = \expectation_{\fullobservation,\action_{j\neq i}\sim\replayBuffer} \big[\nabla_{\param_i}\detpol_i(\action_i|\observation_i)
\nabla_{\action_i} 
Q_i^\policySet(\fullobservation,\action_1,...,\action_N)|_{{\action_i}=\detpol_i(\observation_i)}\big] \,.
\end{split}
\label{eq:maddpg:polgrad}
\end{equation}
In this work we denote the observation received at runtime by agent $i$ as $\observation_i$, and the full state information as $\fullobservation$, from which the observations $o_i$ are derived.
The replay buffer $\replayBuffer$ here contains transitions $(\fullobservation,\action_1,...,\action_N,\reward_1,...,\reward_N,\fullobservation')$ of all agents. 

\section{Overestimation Bias in a Centralized Critic}
\label{sec:overestimation}
\begin{wrapfigure}[18]{r}{0.5\textwidth}
	\includegraphics[trim=0 0 0.4cm 2cm,width=\linewidth]{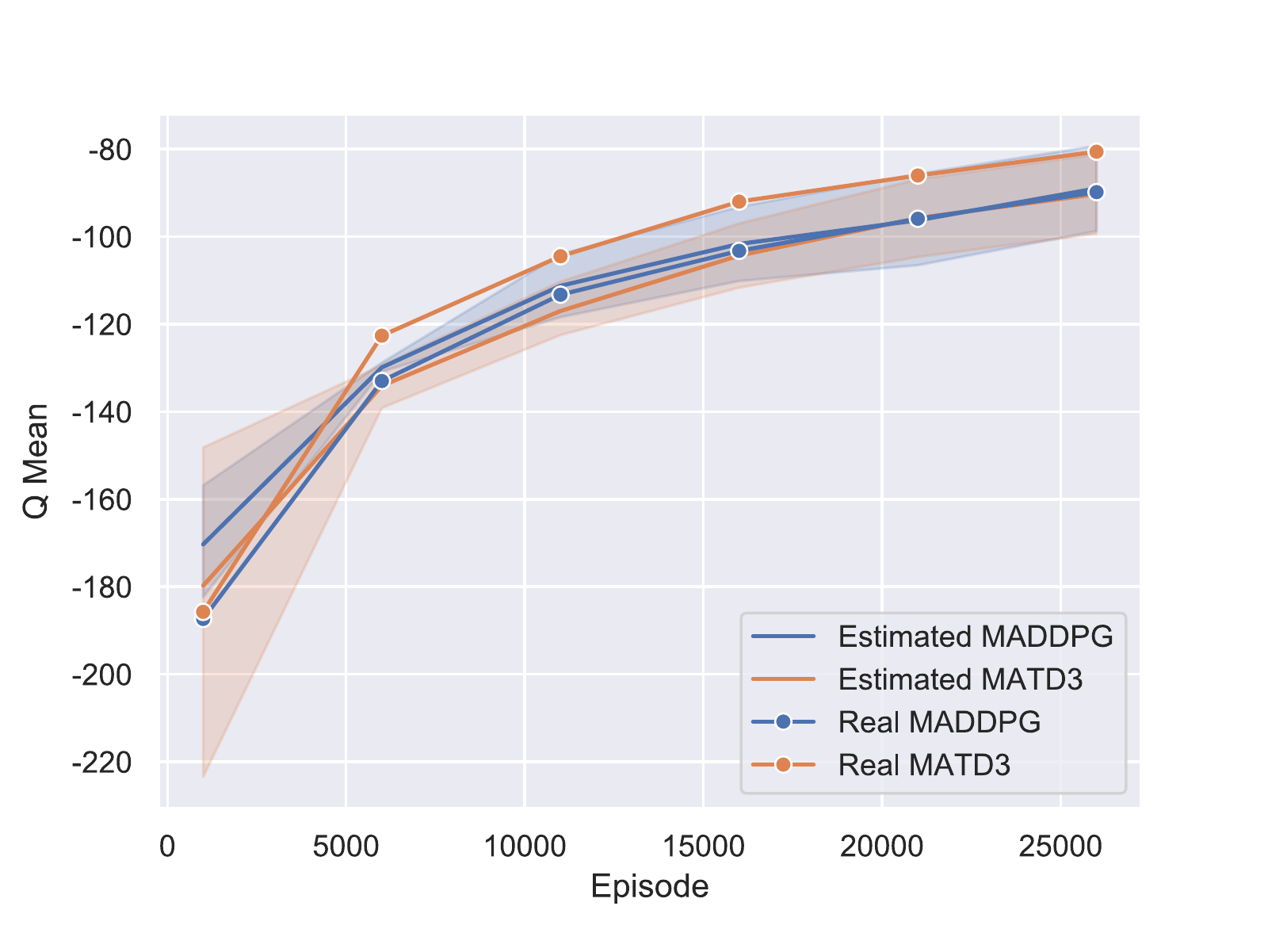}
	\centering
	\caption{Empirical evaluation of overestimation in MARL. The $Q$-values estimated by the $Q$-network and the true $Q$-values are shown. The results are averaged across 5 runs and 95 \% CIs of the mean are shown for the estimated values. We can see, that MADDPG overestimates the $Q$-values, while MATD3 underestimates them and achieves higher real values.}
	\label{fig:overestimation} 	
\end{wrapfigure}
Motivated by related work \cite{Fujimoto2018,VanHasselt2010,VanHasselt2015}, that found an overestimation bias in other methods, we investigate whether this effect persists in the multi-agent domain on the example of the popular \gls*{MADDPG} approach.

As we are using deterministic policies, we can, in the short-term, approximate the environment as stationary from the view-point of each agent, so that the we can regard the transition probability as $\transProb(\state'|\state,\action_1,...,\action_N) \approx \transProb(\state'|\state,\action_i)$.

Under this assumption, we can approximate the value of our centralized critic as $Q_i^\pi(\fullobservation, \action_1,...,\action_N) \approx Q_i^\pi(\fullobservation, \action_i)$, reducing the setting to the one regarded in \cite{Fujimoto2018}, in which they have shown that overestimation occurs in \gls*{DDPG}.
\paragraph{Empirical Evaluation:}

To test whether this overestimation also appears in practice, we evaluated \gls*{MADDPG} on the "Cooperative Navigation" task as outlined in \cite{Lowe2017}.
We increased the number of time-steps per episode to 200 and determine the true and estimated $Q$-values by sampling states and actions from the replay buffer, that were saved since the last evaluation time-step.
From those states, we perform 200 rollouts, with 100 steps each, and save the discounted reward.
We then compare the mean of the discounted rewards with the value of the Q-function approximator.
The results are shown in Figure \ref{fig:overestimation}. 
They show that \gls*{MADDPG} tends to overestimate the $Q$-values, especially during earlier episodes.
Looking at single runs, we can see that this overestimation does not always occur, but when it happens it leads to a significantly worse final performance.

It should also be noted that the evaluated domains are deterministic. 
In contrast to that, most real world applications are stochastic.
Stochasticity has been shown to lead to a higher value function overestimation, because it adds to the noise from function approximator errors \cite{VanHasselt2010}.

\section{Multi-Agent TD3}
\label{sec:approach}
Our proposed approach, called \gls*{MATD3}, extends TD3 to the multi-agent domain in a similar manner to the extension of \gls*{DDPG} to \gls*{MADDPG}.
We use the centralized training with decentralized execution setting, in which we assume that during training we have access to the past actions, observations and rewards, as well as policies, of all agents.
We use this information to learn two centralized critics $Q^\policy_{i,\param_{1,2}}(\fullobservation, \action_1,...,\action_N)$ for each agent $i$.
In order to reduce the overestimation bias, we update them with the minimum of both critics: $y_i = \reward_i + \gamma \min_{j=1,2}Q^\policy_{i,\param_{j}}(\fullobservation',\action_1',...,\action_N')$.
This may lead to underestimation, however, this is preferable to overestimation:
In the case of overestimation, actions with an overestimated value are chosen with a higher probability, due to the policy update.
When then updating the critic, the overestimated value of the next action $\action'$ is used $Q^\policy(\fullobservation', \detpol'(\observation'))$, which propagates the error to the update target $y$.
If it is underestimated, the probability of choosing this action is reduced in the policy update. 
It is thus not used to update the $Q$-values and the error does not propagate further.

In addition, we use target policy smoothing, adding clipped Gaussian noise $\epsilon = \mathtt{clip} (\mathcal{N}(0,\sigma),-c,c)$ to the actions of all agents in the critic update: $\action_j' = \detpol_{\param_j'}(\observation_j') + \epsilon$.
This serves as a regularization, based on the assumption that similar actions should have similar values.
The complete target for the critic resolves to
\begin{equation}
y_i = \reward_i + \gamma \min_{j=1,2} Q^\policy_{i,\param_j'}(\fullobservation',\detpol'_1(\observation_1') + \epsilon,...,\detpol'_N(\observation_N') + \epsilon) \,,
\end{equation}
with $\detpol_i'$ being short for $\detpol_{\param_i'}$.
The policies are updated similar to \eqref{eq:maddpg:polgrad}, but using $Q_{i,\param_1}$ instead of $Q_i$.
We also employ delayed policy updates, only updating the target networks $\param_Q', \param_\policy'$ and policies $\pi_i$ after every $d$ critic updates.
This is motivated in \cite{Fujimoto2018} by the need to have an accurate critic before using it to update the policy, thus updating the critic more often.
This is especially crucial in multi-agent domains, as the change of the critic values has to reflect not only small changes in the own policy, but also in the policies other agents.
However, in adversarial settings this is not always beneficial, as it can slow the adaption to the policy of an adversary.
The full algorithm is shown in the Appendix.

\section{Evaluation in Particle Environments}
\label{sec:particle_eval}
\begin{wrapfigure}[15]{r}{0.58\textwidth}
	\centering
	\includegraphics[trim=0 0 0 0.85cm,width=0.8\linewidth]{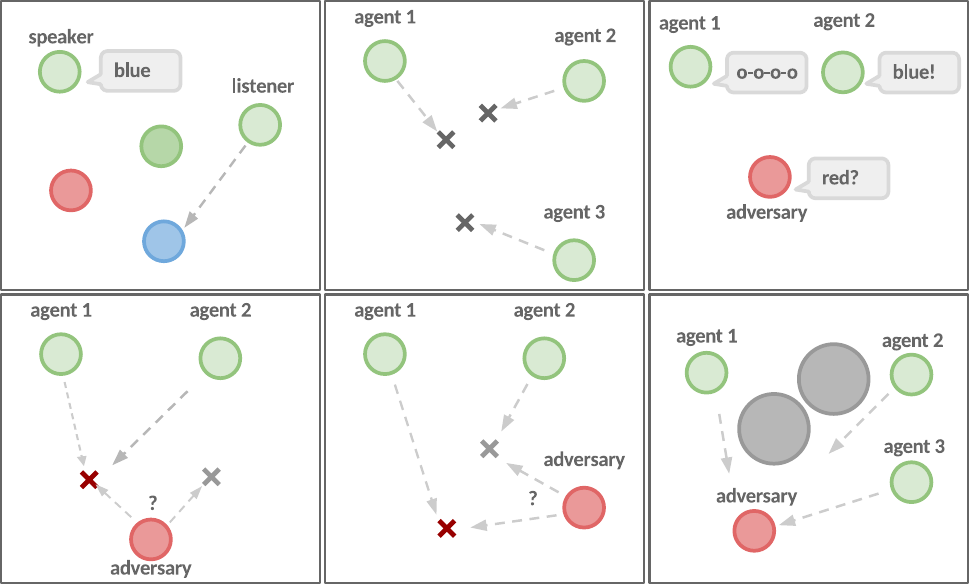}
	\caption{Illustration of the particle environment tasks used in our evaluation. Left to right, top to bottom: "Cooperative Communication", "Cooperative Navigation", "Covert Communication", "Keep-away", "Physical Deception", "Predator-Prey". The figure is based on \protect\cite{Lowe2017}.}
	\label{fig:particle-envs}
\end{wrapfigure}

We evaluate the efficacy of our approach on the particle environments proposed by \cite{Mordatch2017} and used to evaluate \gls*{MADDPG} in \cite{Lowe2017}.  They are shown in Figure \ref{fig:particle-envs}.
The particle environments consist of two-dimensional continuous state spaces, in which agents can exert a force on themselves.
Additionally, the agents may have access to a discrete communication channel to each of the other agents.

The particle environments consist of a set of six tasks:
Two cooperative tasks called "Cooperative Navigation" and "Cooperative Communication", in addition to four adversarial tasks "Covert Communication", "Keep-Away", "Physical Deception" and "Predator-Prey". 
In all of the adversarial tasks, there is a team of "Agents" and a single "Adversary".
Most of the tasks require a team of agents to learn a cooperative strategy, which can deal with most behaviors of the adversarial agent.
An exception is the "Covert Communication" task, in which the adversary has to decode a message the agents are sending to each other. 
In this task a good result can be achieved by quickly changing the communication scheme, without learning a more complex behavior.

\subsection{Results}
\begin{wrapfigure}[15]{r}{0.5\textwidth}
	\includegraphics[trim=0 0 0 2.5cm,width=\linewidth]{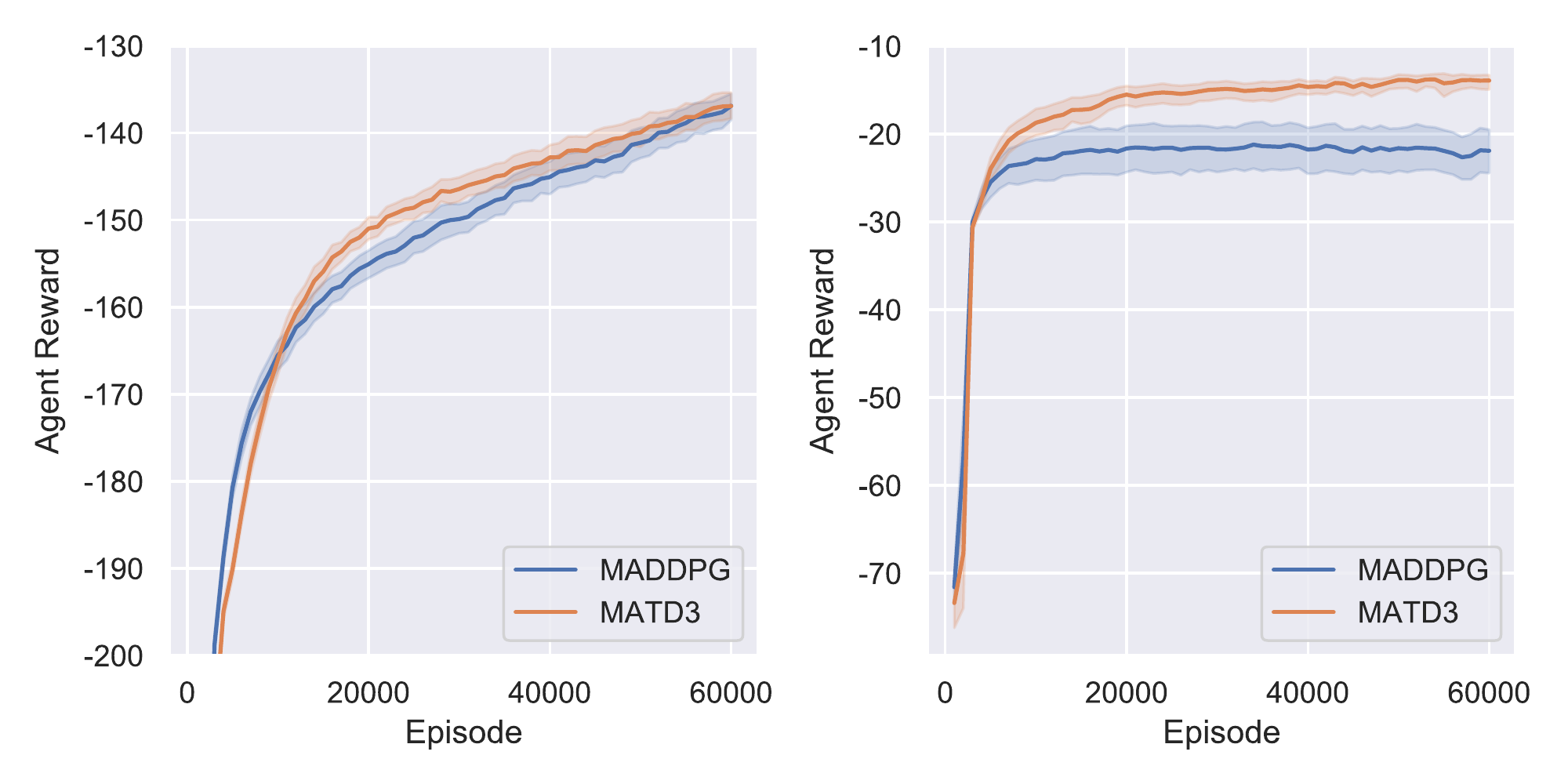}
	\caption{Evaluation in the cooperative domains used in \protect\cite{Lowe2017}, "Cooperative Navigation" (left) and "Cooperative Communication" (right). We can see, that MATD3 significantly outperforms MADDPG. Shown is the mean episodic reward over the last 1000 episodes, shaded areas are the 95 \% confidence intervals of the mean, averaged across 20 trials.}
	\label{fig:cooperative-eval}
\end{wrapfigure}
We implement our approach, named \gls*{MATD3}, and compare its performance to \gls*{MADDPG}.\footnote{The source code will be available after the review period, to ensure anonymity.}

Hyper-parameters are chosen by grid-search over learn rate $ \alpha = [0.01, 0.003, 0.001]$, mini-batch size $b = [256, 1000]$ and policy update frequency $ d = [1,2,3]$. 
The parameters we found to work best are $\alpha = 0.01$, $b = 1000$, $d = 2$.
We use the same set of parameters for all tasks, to ensure a fair comparison.

For \gls*{MADDPG} we use the hyper-parameters and implementation provided in \cite{Lowe2017}, which were tuned for the same tasks.
For both approaches we approximate the Q-functions and policies with \glspl*{MLP} with two hidden layers with 64 units each.
As activation function we use the \gls*{ReLU} function, and as optimizer we use use Adam \cite{Kingma2014} in all experiments, as well as the Gumbel-Softmax estimator \cite{Jang2017}.
\paragraph{Cooperative Environments}
The results in the cooperative tasks are shown in Figure \ref{fig:cooperative-eval}. 
On the cooperative navigation task both achieve a similar final performance, while MATD3 learns a better policy significantly faster.
On the cooperative communication task \gls*{MATD3} achieves a significantly better final performance.

\paragraph{Competitive Environments}
\begin{figure}[t]
	
	\includegraphics[width=0.98\textwidth]{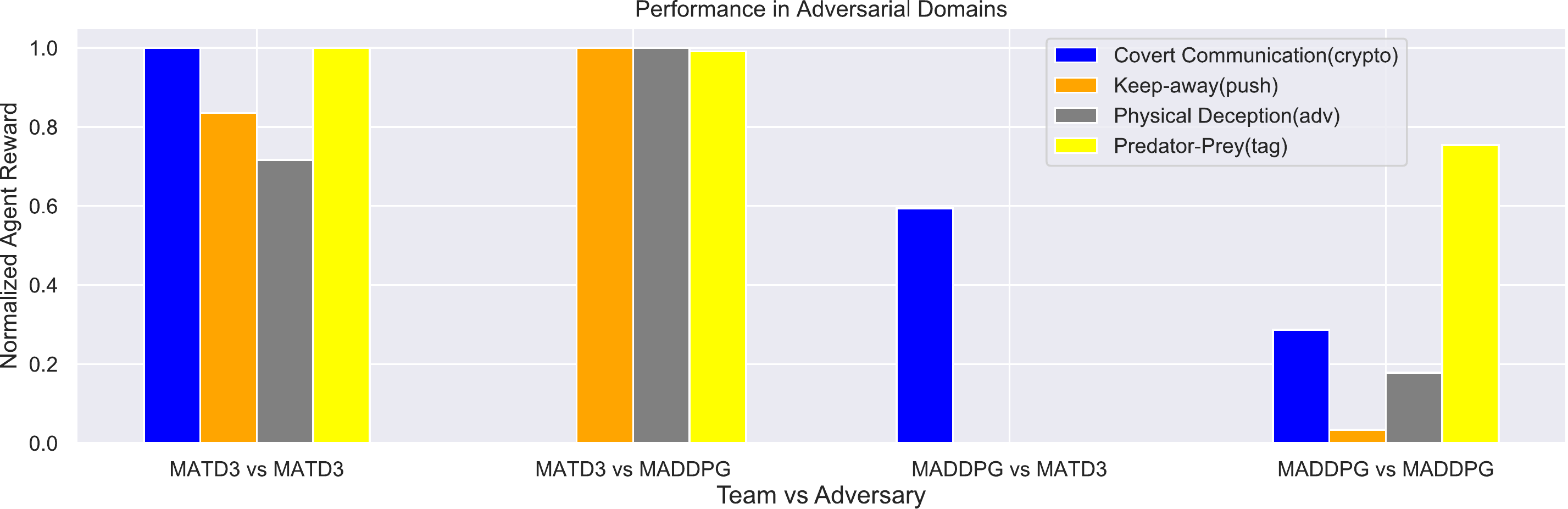}
	\centering
	\caption{Evaluation in the adversarial domains (Team vs Single Agent). 
		Shown is the 0-1 normalized final reward of the team, in all combinations of MATD3 and MADDPG, averaged across 20 trials each.
		In the domains where it is necessary to learn a stable winning strategy ("Keep away", "Physical-Deception", "Predator-Prey") MATD3 outperforms MADDPG in the direct comparison.
		However, in the "Covert Communication" domain, where quick adaption to the policy of the adversary is advantageous, MADDPG outperforms MATD3.
	}
	\label{fig:competitive-eval}
	\vspace{-0.2cm}
\end{figure}
Results in the competitive environments are shown in Figure \ref{fig:competitive-eval}, as 0-1 normalized, final rewards, averaged across 20 trials each.
They show that, in direct comparison, MATD3 outperforms MADDPG in three out of four environments.

\begin{figure}
	\begin{minipage}[t]{0.48\textwidth}
		\centering
		\includegraphics[width=\columnwidth]{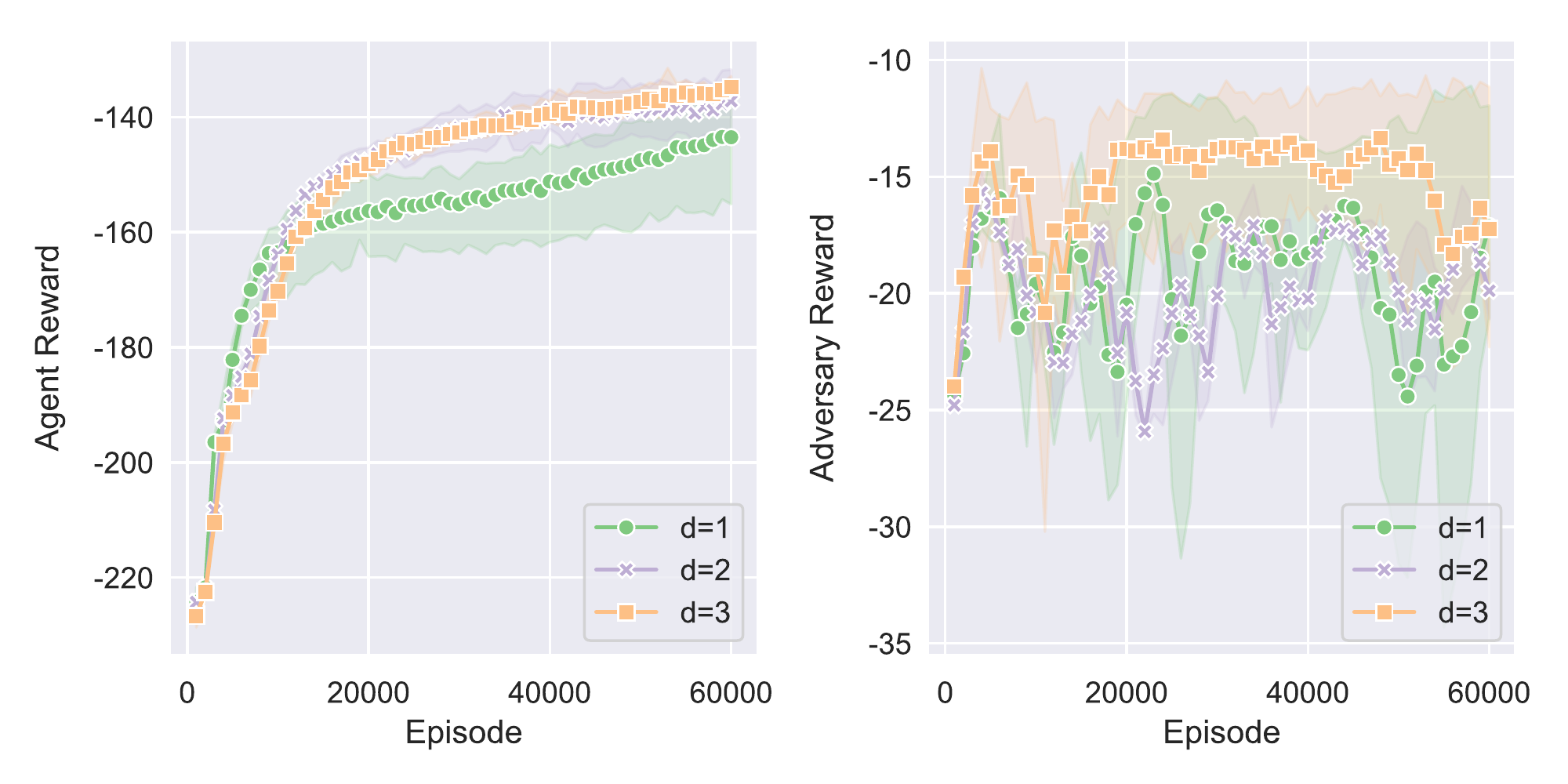}
		\caption{Effect of the policy update rate $d$ on performance in "Cooperative Navigation" (left) and "Covert Communication" (right). 
			We can see, that a less frequent policy update is beneficial in the cooperative task, while in the adversarial task it leads to a better performance of the Adversary, i.e., it being better at decrypting the communication of the agent team.}
		\label{fig:pol_rate}
	\end{minipage}
	\hfill
	\begin{minipage}[t]{0.48\textwidth}
		\includegraphics[trim= 0 0 0.9cm 0.9cm, width=\linewidth]{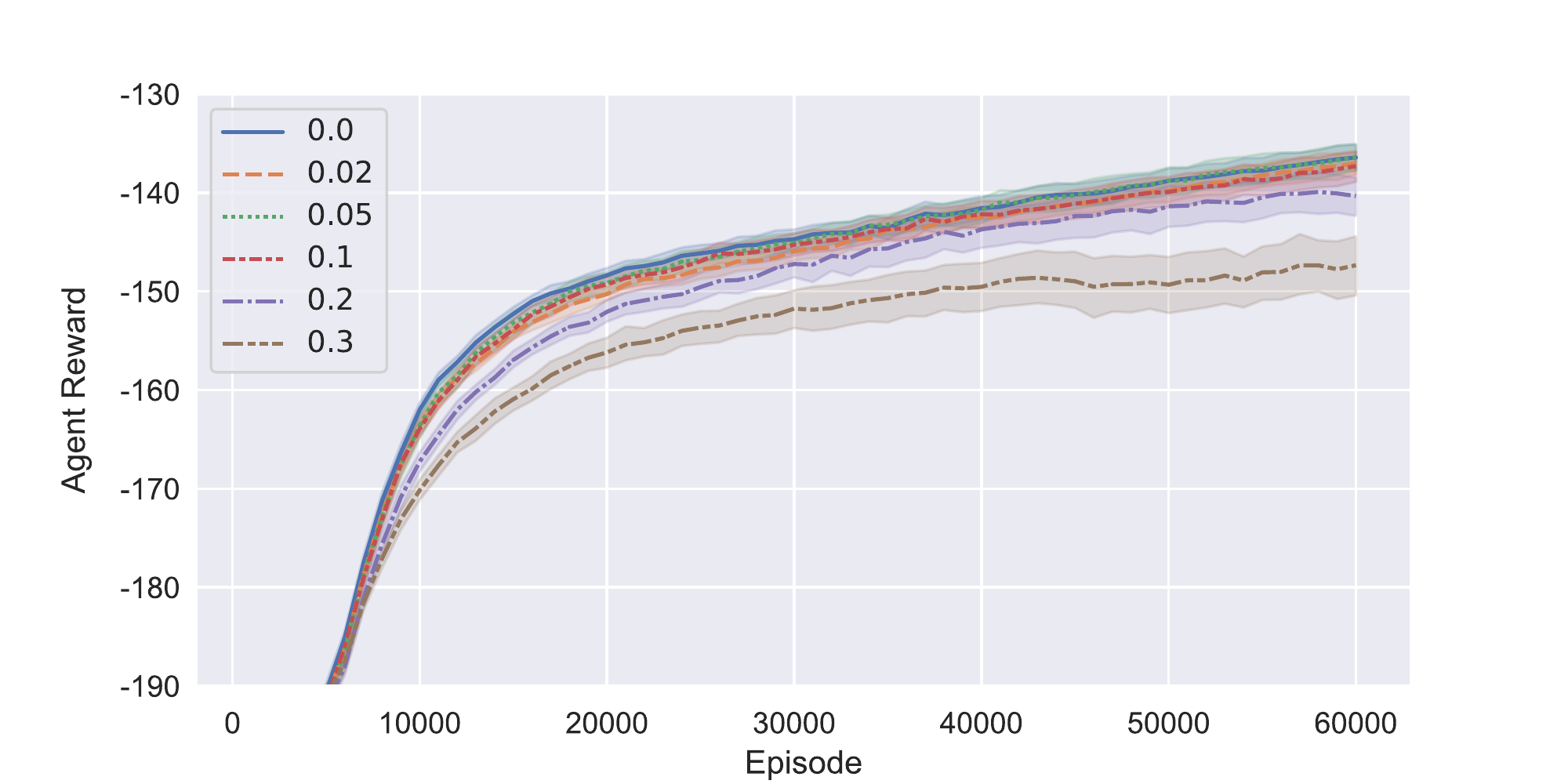}
		\centering
		\caption{Evaluation of target policy smoothing on the "Cooperative Navigation" task. Shown is the mean reward for different values for $\epsilon$, averaged across 10 runs each. Note the zoomed in axis. In our evaluation we did not find a significant advantage of target policy smoothing.}
		\label{fig:target_policy_smoothing}
		\vspace{-0.5cm}
	\end{minipage}
\end{figure}

In the task where \gls*{MADDPG} significantly outperforms our proposed approach, 'Covert Communication', a team of agents has to learn a communication strategy, that the single adversary has to decode.
Due to the delayed policy updates, MATD3 is slower at adapting to its opponent's behavior, thus being outperformed by MADDPG.
In the other tasks a consistent winning strategy, that can beat all behaviors of the single agent, can be learned, at which our proposed approach succeeds.
\paragraph{Delayed Policy Updates}
Delayed policy updates are intended to ensure a sufficiently converged critic before using it to update the policy.
To investigate their effect in \gls*{MARL}, we evaluated different policy update rates and show the results in Figure \ref{fig:pol_rate}.
Less frequent policy updates showed to be beneficial in all tasks, leading to a lower variance in results and a higher final performance, with the exception being the "Covert Communication" task. 
In this task, the team of agents is made slower at changing it's communication policy, leading to the adversary being better at decrypting it.
\paragraph{Target Policy Smoothing}
We evaluate the effect of target policy smoothing in Figure \ref{fig:target_policy_smoothing} for different levels of added noise $\epsilon$.
We do not find target policy smoothing to improve the performance unlike in \cite{Fujimoto2018}.
We assume that this is due to the policies of the other agents used in the critic target being updated frequently, and thus implicitly introducing a similar randomness.

We also evaluated the relative overestimation for different numbers of agents, but did not find a significant difference.

\vspace{-1.5mm}
\section{Learning Fully Decentralized Controllers for Robotic Systems}
\vspace{-1.5mm}
\label{sec:robot-decentral}
The particle environments regarded in the previous section are mostly fully observable, with the actions of one agent often not strongly affecting the other agents.
Thus, in many tasks, a high reward can be achieved without much cooperation.
We therefore also evaluate our approach in a new, challenging setting: Learning decentralized controllers for robotic systems.
In high-dimensional robotic tasks decentralized control has been shown to be an effective method, enabling efficient locomotion \cite{Whitman2016}. 
It commonly functions by virtually subdividing the robot into multiple parts. 
These parts are then coordinated by a central controller.
Additionally, in many cases this kind of decentralization is required, for example due to band-width limitations or the structure of the robot.

We propose to eliminate the need for a centralized or shared control policy by regarding the robot as a multi-agent system.
Furthermore, this does not require the additional model knowledge to design a decentralized controller.
We thus partition the robotic system into multiple agents, which only have access to partial information about the state of the other agents. 
The agents learn to coordinate their actions based on the reward signal they receive and the centralized critic.

The reduction to a partial observation for each agent leads to the task becoming a partially observable stochastic game \cite{Hansen2004}, that requires a large degree of cooperation between agents.
We therefore do not aim to outperform the current state-of-the-art methods proposed for single-agent continuous tasks, as they have access to the full-state, but see this as a new, challenging task for \gls*{MARL}.

\vspace{-1.5mm}
\subsection{Decomposition to Multiple Agents}
\vspace{-1.5mm}
We evaluate the functionality of our proposed approach on the OpenAI Gym \cite{gym} "Ant-v2" task.
The ant consists of a spherical torso and four legs.
The legs each have two actuated joints, one at the attachment to the torso and one at the knee.
The state information provided in the standard Gym task consists of all joint positions and angular velocities, position and velocity of the torso as well as contact forces with the floor.
This is then used to generate a reward consisting of the distance traveled in a set direction and a cost term for actuator force and contact with the surface.
Additionally, a positive reward is obtained for every time-step that the agent does not reach a terminal position, which occurs when the torso falls below a threshold height.

\begin{figure}[t]
	\begin{minipage}[t]{0.4\textwidth}
		\centering
		\includegraphics[width=\columnwidth]{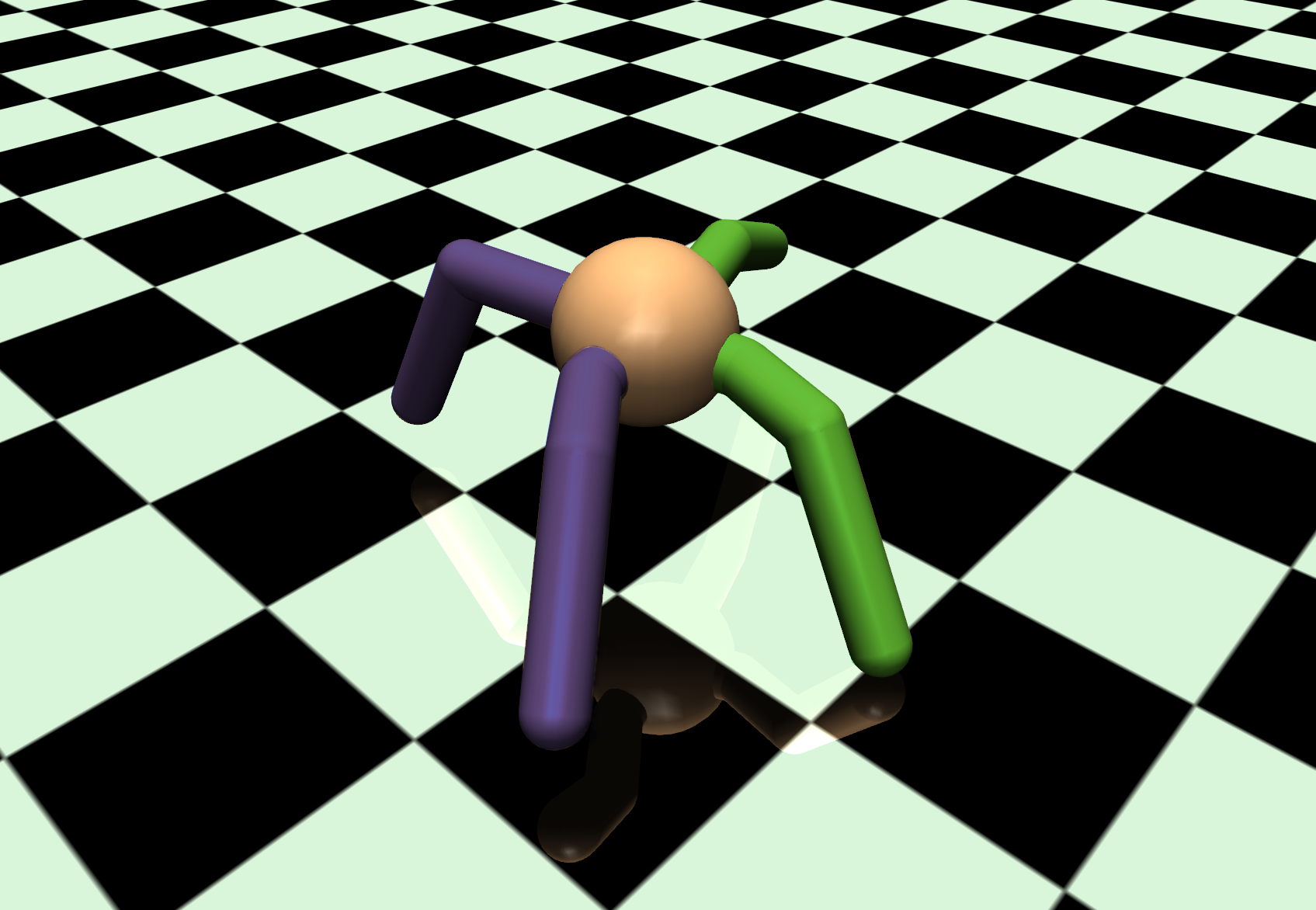}
		\caption{The "Ant-v2" task, split into two agents, visualized as the green and blue part. The observation of each agent consists of full information of its side, but only includes the joint positions of the other side, without acting forces or velocities.}
		\label{fig:antsplit}
	\end{minipage}
	\hfill
	\begin{minipage}[t]{0.57\textwidth}
		\centering
		\includegraphics[trim=0 0 1.5cm 0, width=0.96\columnwidth]{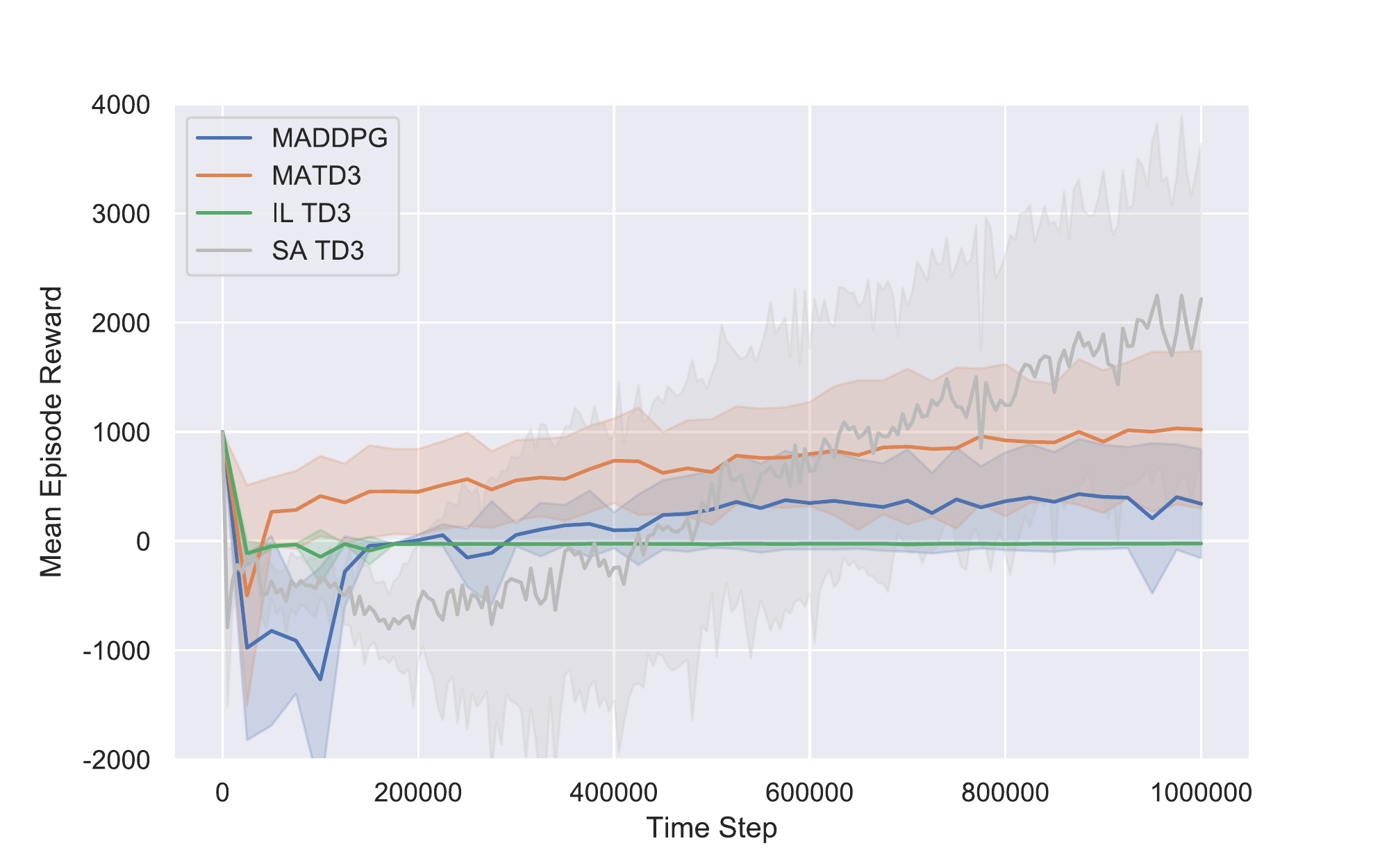}
		\caption{Performance on the "Ant-v2" task. Mean reward of deterministic evaluation episodes, averaged over 6 seeds per approach. The shaded area is a standard deviation. Shown are \gls*{MATD3}, \gls*{MADDPG} and independent learner (IL) TD3 on the decentralized task. For comparison we also show the performance of a single agent (SA) using \gls*{TD3}, with full state information. Note that this is a significantly less difficult setting.}
		\label{fig:antresults}
	\end{minipage}
\end{figure}

We separate the ant into two halves, as shown in Figure \ref{fig:antsplit}.
The action-space of one agent comprises the two left legs, and the action-space of the other consists of the other two legs.
As observation each agent receives all information about their respective legs, that is provided in the "Ant-v2" task - position, angular velocity and external forces - but only the position of the other legs.
In addition, both agents receive the location and velocity of the torso of the ant.

\subsection{Implementation}
We implemented our approach, MATD3, for the ant experiments, along with a MADDPG and independent learner (IL) TD3 version. To our knowledge this is the first investigation of the efficacy of MADDPG in high-dimensional, continuous action spaces.

The hyperparameters for \gls*{MATD3} and \gls*{MADDPG} were selected by grid-search over learn rate $\alpha = [0.01, 0.003, 0.001]$, batch-size $ b = [100, 300]$ and $ \tau = [0.005, 0.01]$.
The hyper-parameters we found to be working best for both approaches are $\alpha = 0.001$, $b = 100$ as batch-size and $\tau = 0.01$, and $d=2$ for \gls*{MATD3}.
For single agent (SA) \gls*{TD3} and IL \gls*{TD3} we are using the hyper-parameters suggested in the original paper \cite{Fujimoto2018}.

For all Q-functions and policies we use \glspl*{MLP} with two hidden layers with 400 and 300 units respectively. 
As activation function we use \gls*{ReLU}, except at the output, which uses a sigmoid activation. 
The output is then scaled linearly to the range of the respective action space.
\vspace{-1mm}
\subsection{Results}
\label{sec:robot-eval}

The results of our trials are shown in Figure \ref{fig:antresults}. 
They show that \gls*{MATD3} performs better than \gls*{MADDPG}.
Both approaches usually first find a policy that remains in place, however, our approach, unlike \gls*{MADDPG}, usually recovers from this local optimum and achieves locomotion in the required direction.
As a baseline we also show the performance of two independent learner (IL) \gls*{TD3} agents, which receive the same observation as the \gls*{MATD3} and \gls*{MADDPG} agents, but do not use a centralized critic.
They failed to learn a successful policy in all trials.

Finally, we show the performance of SA TD3 in the standard, fully observable "Ant-v2" task.
Unsurprisingly, due to the SA task being significantly easier, the final performance of SA TD3 outperforms all MA approaches.
However it only starts to do so after a high number of time-steps, showing promise for further work.

\vspace{-1.5mm}
\section{Related Work}
\vspace{-1.5mm}
Regarding the improvement of \gls*{MADDPG}, Minimax Multi-Agent DDPG (M3DDPG) \cite{Li2019} has to be noted.
They approximate a minimax training objective by adding adversarial perturbations to the actions of other agents when updating the critic and policy.
However, their improvements are limited to adversarial tasks, while we also address cooperative ones.
Additionally, it should be possible to combine their approach with ours.
An approach that aims to reduce overestimation bias in \gls*{MARL} by using Double Deep Q Networks (DDQN) is presented in \cite{Simoes2017}. 
However, they do not investigate whether overestimation does indeed occur in \gls*{MARL} and if their approach reduces it.
Further, their work focuses on discrete state and action spaces in a grid-world, while our work focuses on more complex, continuous domains.
Furthermore, DDQN has been shown to not be effective in the actor-critic setting \cite{Fujimoto2018}.

In the field of robotics, learning decentralized controllers via \gls*{RL} has been studied by \cite{Busoniu2006}. 
Instead of using a centralized critic, they use independent critics which are augmented by certain additional observations of the other agents. 
In addition, they use value iteration to learn the policies, which does not scale to high-dimensional tasks, and only study comparatively simple tasks.
\vspace{-1.5mm}
\glsresetall
\section{Conclusion and Future Work}
\vspace{-1.5mm}
We have shown, that overestimation occurs in multi-agent domains and significantly hinders convergence.
We used this finding to propose a new approach for \gls*{MARL}, called multi-agent TD3 (MATD3).
It is based on the decentralized execution, centralized training setting and addresses the overestimation bias by using double centralized critics.
We have shown that our proposed approach significantly outperforms \acrshort*{MADDPG} on most of the particle-domain tasks.
In addition, we propose a new method of learning decentralized controllers for robotic tasks, by regarding them as multi-agent systems and using methods from \gls*{MARL}.
We showed that we can use this approach to learn decentralized policies for the popular "Ant" task, and that our proposed approach also outperforms \acrshort*{MADDPG} in this domain.

For future work, we plan to investigate a hybrid approach, that combines the initial benefits of \gls*{MATD3} with the later performance of TD3.

\appendix
\begin{algorithm*}
	Initialize replay buffer $\replayBuffer$ and network parameters\;
	\For{$t = 0$ to $T_\text{max}$}{
		Select actions $\action_i \sim \detpol_{i}(\observation_i) + \epsilon$ \;
		Execute actions $(\action_1,...,\action_N)$, observe $\reward_i$, $\fullobservation'$\;
		Store transition $(\fullobservation,\action_1,...,\action_N,\reward_1,...,\reward_N,\fullobservation')$ in $\replayBuffer$ \;
		\Let{$\fullobservation$}{$\fullobservation'$}\;
		\For{agent $i=1$ to $N$}{
			Sample a random minibatch of $S$ samples $(\fullobservation^b,\action^b,\reward^b,\fullobservation'^b)$ from $\replayBuffer$\;
			$y^b \gets r_i^b + \decay \min_{j=1,2} Q_{i,j}^{\detpol'}(\fullobservation'^b,\action_1,...,\action_N)\mid
			_{\action_k = \detpol_k'(\observation'^b_k) + \epsilon}$ \;
			Minimize Q-function loss for both critics $j=1,2$\;
			$\loss({\param_j}) = \frac{1}{S}\sum_b(Q_{i,j}^\detpol(\fullobservation^b,\action_1^b,...,\action_N^b) -y^b)^2$\;
			\If{$t$ mod $d= 0$}{
				Update policy $\detpol_{i}$ with gradient
				$$\nabla_{\param_{\detpol,i}}  J \approx
				\frac{1}{S}\sum_{b}
				\nabla_\param\detpol_{\param_{\detpol,i}}(\observation^b_i)\nabla_{\action_i} Q_{i,1}^\mu(\fullobservation^b,\action^b_1,...,\detpol_{\param_{\detpol,i}}(\observation_i),...\action^b_N)
				$$
				Update the target networks $\param' \gets \tau \param + (1 - \tau) \param'$ \;
			}
		}
	}
	\caption{Multi-Agent TD3}
	\label{alg:matd3}
\end{algorithm*}


\end{document}